\pdfoutput=1
\documentclass[11pt]{article}

\usepackage[final]{acl}

\usepackage{times}
\usepackage{booktabs}
\usepackage{tabularx}
\usepackage{latexsym}

\usepackage[T1]{fontenc}
\usepackage[utf8]{inputenc}

\usepackage{pdfpages}

\usepackage{graphicx}
\usepackage{microtype}

\usepackage{inconsolata}
\usepackage{amsmath} 

\usepackage{listings}
\usepackage{lipsum}

\usepackage{xcolor}
\definecolor{myblue}{RGB}{123, 194, 76}

\newcolumntype{b}{X}
\newcolumntype{s}{>{\hsize=.15\hsize}X}
\newcolumntype{m}{>{\hsize=.3\hsize}X}

\title{EDEN: \textbf{E}mpathetic \textbf{D}ialogues for \textbf{E}nglish learning}

\author{Li Siyan, Teresa Shao, Zhou Yu, Julia Hirschberg\\
Department of Computer Science\\
Columbia University\\
\texttt{\{siyan.li,ts3488,zy2461,jbh2019\}@columbia.edu}}

\begin{document}
\maketitle
\begin{abstract}
% Dialogue systems have been used as conversational partners.., howveer, very few studies have studied whether these systems can improve learning outcomes.

% focusing on how the students' perception of the chatbots' ability to support them emotionally correlates with how their
Dialogue systems have been used as conversation partners in English learning, but few have studied whether these systems improve learning outcomes.
% We study the relationship between an English-teaching chatbot's perceived affective support and changes in student grit.
Student passion and perseverance, or \textit{grit}, has been associated with language learning success. Recent work establishes that as students perceive their English teachers to be more supportive, their grit improves.
Hypothesizing that the same pattern applies to English-teaching chatbots, we create EDEN, a robust open-domain chatbot for spoken conversation practice that provides empathetic feedback. To construct EDEN, we first train a specialized spoken utterance grammar correction model and a high-quality social chit-chat conversation model.
We then conduct a preliminary user study with a variety of strategies for empathetic feedback. Our experiment suggests that using adaptive empathetic feedback leads to higher \textit{perceived affective support}. Furthermore, elements of perceived affective support positively correlate with student grit.
% As dialogue systems gain prominence in English language education, we must identify means of improving student success in these systems. One approach is increasing \emph{grit}, which is perseverance and passion in learning. 
% %JH: probably good to define it here too since i don't think many people know what this means
% Grit is an important protective factor against burnout. We study the relationship between an English-teaching chatbot's perceived affective support and changes in user grit. We additionally explore the effect of different empathetic feedback on grit. Our experiment suggests that using adaptive empathetic feedback leads to higher perceived affective support, which correlates with positive changes in grit. Our results pave the way for English-teaching systems that mitigate burnout and improve mental health. 
\end{abstract}

% retain users over extended period of time in order to maximize their learning.

% 1) There are English-teaching chatbots
% 2) 

% Language learners must interact with their tutors or language-teaching chatbots persistently to maximize their learning

% The increasing popularity of dialogue systems has led to the creation of English-teaching chatbots \cite{ayedoun2020toward,ayedoun2015conversational,yang2022implementation,kohnke2023l2}. Since language learning requires persistent practice for an extended period of time, learners would need to interact with these chatbots consistently for these tools to be useful. One factor determining whether a second-language (L2) learner would be willing to practice frequently is L2 grit, defined as passion and perseverance for second-language learning \cite{teimouri2022l2}.

% L2 grit has been widely studied in language learning. 

\section{Introduction}
We study chatbots that teach languages like English~\cite{ayedoun2020toward,ayedoun2015conversational,yang2022implementation,kohnke2023l2}, and in particular how they can improve student persistence in learning a second language. 
In the language learning literature, this is referred to as \textit{L2 grit} \cite{teimouri2022l2}.
High L2 grit is crucial for student well-being and success. For example, L2 grit correlates strongly with increased learning enjoyment \cite{elahi2021longitudinal}, negatively predicts foreign language anxiety and burnout \cite{li2021classroom,wu2023teachers}, and indirectly predicts L2 achievement \cite{khajavy2022contribution}. While L2 grit is often framed as a personality trait, changing an individual's grit is possible \cite{hwang2021enhancing,tang2019building,pueschel2018achieving}. As a result, a promising direction is to study grit-improving interventions.

\begin{figure}
    \centering
    \includegraphics[scale=0.7]{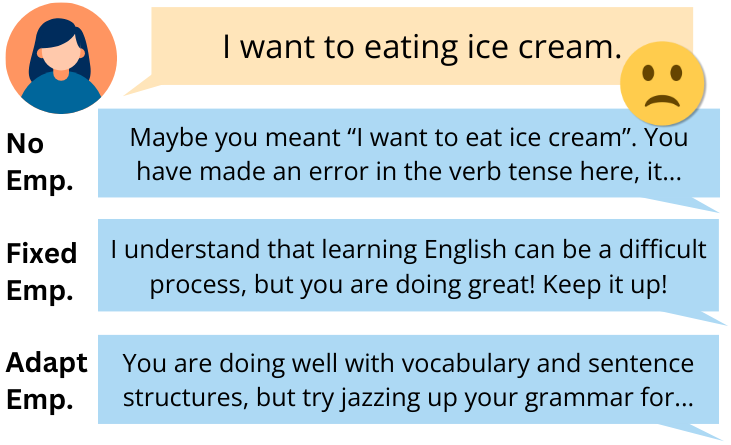}
    \caption{The three different empathetic feedback strategies in our experiments. This is a special case where the input is grammatically incorrect, so the No Empathetic Feedback condition would provide corrections.}
    \label{fig:enter-label}
\end{figure}

\citet{wu2023teachers} examines how teachers' perceived affective support (PAS), i.e. how supportive the students perceive their teachers to be \cite{sakiz2007does}, influences student L2 grit in Chinese ESL medium-level learners in a college-level English class. The authors discover a strong predictive relationship between higher teacher PAS and higher student L2 grit. This further encourages teachers to exhibit warmth and empathy to improve their PAS. 
The study does not establish whether higher teacher PAS improves L2 grit, but we posit that it is possible. 

We seek to determine whether the relationship between PAS and L2 grit extends from human teachers to English-teaching dialogue systems. That is, \textbf{does higher \textit{chatbot} perceived affective support also increase student L2 grit?} Since perceived affective support correlates positively with empathy, an empathetic English-teaching chatbot should allow us to study this relationship. However, there is little work on incorporating empathy into open-domain English-teaching chatbots \cite{zhai2022systematic}.

% Previous paper is a demo, but we expand the system to be more robust, downplay my previous paper A LOT, first attempt at building a demo, however it is a demo paper, that does not include concrete user study that demonstrates the results

\begin{figure*}
    \centering
    \includegraphics[scale=0.7]{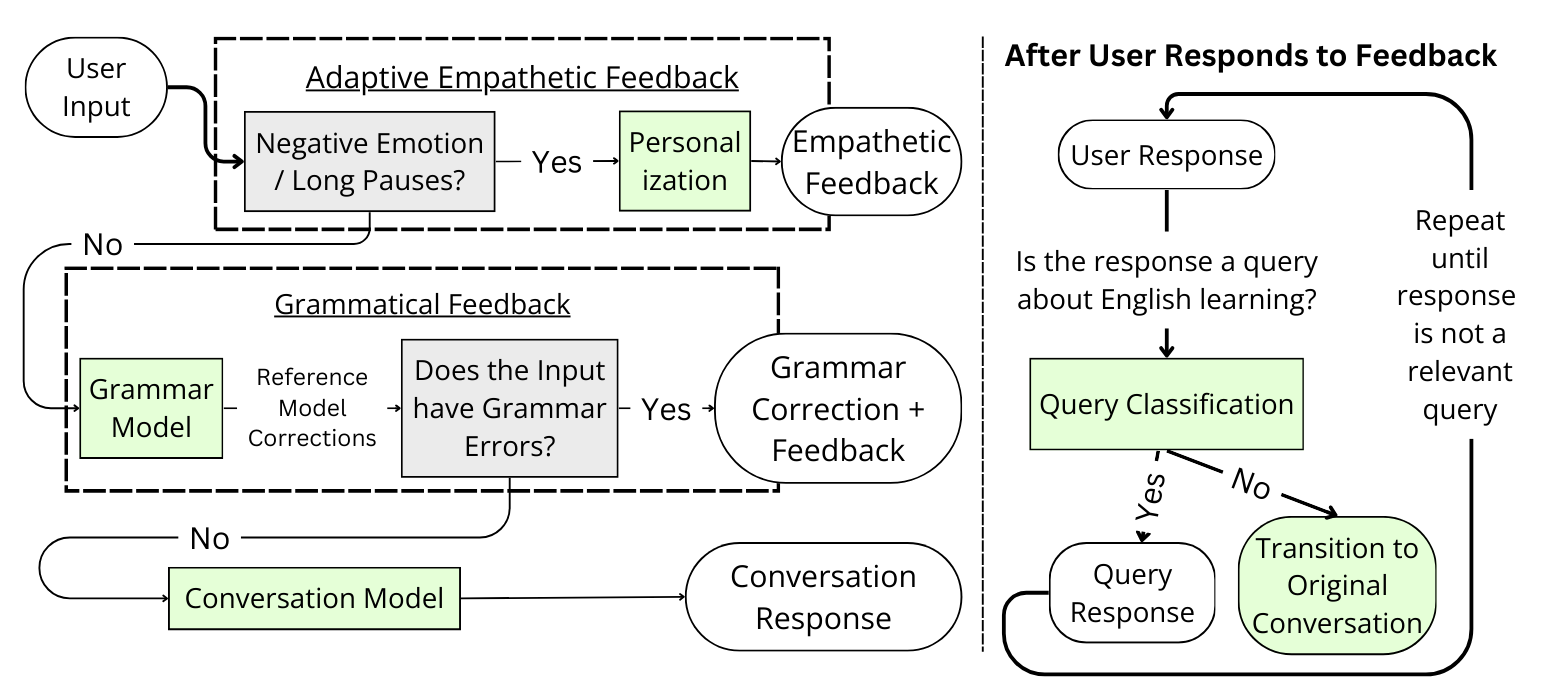}
    \caption{An overview of EDEN's architecture. We highlight several additions and improvements compared to the design by \citet{siyan2024using} in green. }
    \label{fig:changes}
\end{figure*}

To bridge this gap, we construct EDEN (\textbf{E}mpathetic \textbf{D}ialogues for \textbf{EN}glish learning), which is a high-quality and robust dialogue system capable of empathetic and grammatical feedback. 
To strengthen components of EDEN for educational spoken dialogue, we tailor our grammatical feedback for spoken utterances, build a conversation model for open-domain chitchat across multiple topics, and introduce personalization to cater to user preferences. 
We encourage others to expand upon and customize EDEN for research within and outside of language education.\footnote{We release all data, code, and model checkpoints under an open-source license here: \url{https://github.com/siyan-sylvia-li/EDEN}}

% Their initial study shows that native Mandarin speakers with intermediate English proficiency perceive the chatbot to have medium affective support. 
% While the authors introduce a novel pipeline for incorporating empathy in chatbot responses, their system (i) can only hold conversations on one topic, i.e. movie discussions, (ii) cannot adapt to different learner preferences, which the authors found leads to sub-optimal user experience, (iii) does not account for the spoken nature of the task in offering grammatical feedback, and (iv) fails to detect conclusive results due to a small-scale user study. 
% In this work, we first significantly improve this system by building a personalizable chatbot that can hold conversations across seven primary topic areas and that tailors grammatical feedback for spoken utterances. We then use our new chatbot system to conduct our study.
% To our knowledge, this chatbot is the first English-teaching chatbot to incorporate speech-based emotion recognition and subsequent empathetic intervention \cite{mercer2016seeing,rashkin2018towards}.
%%JH: i'd add a link to a paper on empathy here
% Therefore, we decide to use this chatbot to test the correlation between chatbot PAS and user L2-grit and to explore strategies for chatbot PAS enhancement.

Using EDEN, we conduct a preliminary user study on how empathetic feedback mechanisms influence the chatbot's perceived affective support and user L2 grit changes. Our results suggest that the adaptive empathetic feedback strategy is the most successful in inducing high perceived affective support. This could be due to the specificity of the adaptive mechanism making users feel more thoughtfully attended to. Additionally, we discover that certain components of chatbot perceived affective support predict positive changes in L2 grit, which aligns with our hypotheses.

\section{Related Work}
% Include diane littman's work for physics, we speculate that STEM learning should contribute to language learning

Empathetic chatbots have been applied to counseling \cite{devault2014simsensei,trappey2022development}, medical assistance \cite{daher2020empathic}, motivation for weight management \cite{rahmanti2022slimme}, customer service \cite{xu2017new}, or for social and communicative needs \cite{de2020effectiveness,jiang2022chatbot}. However, there is little work on integrating empathy into second-language education using current-day language technologies.
% In spite of this variety, one should note that language education requires a different form of empathy. Language learners experiencing negative emotions may not always describe their distress.
%JH: do you want to say what that is, perhaps briefly here?

There have been affective English educational conversation systems. \citet{ayedoun2020toward}, improving upon \citet{ayedoun2015conversational}, presents a multimodal agent for improving L2 learners' willingness to communicate. The agent carries out a pre-scripted dialogue and adopts different communicative strategies and affective backchannels to reduce learner anxiety. \citet{shi2020design} creates an empathetic spoken chatbot for pronunciation correction using an ontology. \citet{lee-etal-2023-peep} trains a real-world situational chatbot capable of providing feedback. \citet{park-etal-2022-freetalky} incorporates persona-based conversation capabilities into a humanoid robot to make language practice easier for anxiety-prone individuals. Nonetheless, none of these chatbots account for student emotions. The learner-emotion-aware systems, on the other hand, are often not conversational \cite{lin2015perspective,wu2022affective,santos2016toward}. 
% \citet{lin2015perspective} utilizes facial emotion recognition to adjust learning material difficulty if students exhibit negative emotions. The system contains a non-interactive teaching assistant avatar that mirrors learner-expressed emotions. Similarly, \citet{wu2022affective} uses the number of times the learner clicks on different emotion-related words to estimate learner affect, and an agent responds accordingly. Although there are interactions present, the format is not free-form dialogue, and the emotion recognition approach is not seamless. \citet{santos2016toward} constructs a ubiquitous learning classroom to record and react to L2 learners' physiological signals associated with negative affective states. The system engages by providing educational recommendations to learners and not direct conversation. 

Empathetic strategies have been employed in other forms of learning. Affective AutoTutor \cite{d2013autotutor} responds with emotional statements to regulate negative student emotions in physics tutoring. \citet{litman2014evaluating} modifies a spoken physics education system \cite{litman2004itspoke} that adapts to user affective states to identify and respond to real-time user disengagement. Other affective tutoring systems often offer hints to resolve student frustration \cite{hasan2020transition,fwa2018architectural,lin2014influence}. We postulate that these approaches may be proven effective for language learning as well. 

Prior work has explored the relationship between chatbot usage and L2 learner experience. \citet{han2021analysis} reports Korean EFL learners experiencing reduced language anxiety and enhanced English-learning interests when using English chatbots. AI-mediated discussions have also been more effective than face-to-face discussions for increasing L2 learners' willingness to communicate \cite{fathi2024improving}. However, there is no systematic study on L2 grit changes in the chatbot context.

% There is also a shortage of systematic studies on the effect of English-teaching chatbots on L2 developmental factors.

% \Sylvia{L2 feedback, criticism sandwich}

% \Sylvia{Motivation for our different criteria}

% \subsection{Empathetic Systems for L2 Learning}

% \begin{figure}
%     \centering
%     \includegraphics[scale=0.15]{latex/place.jpg}
%     \caption{Caption}
%     \label{fig:enter-label}
% \end{figure}

\section{Chatbot Design}
While using dialogue systems for English education has become popular, there are only a few fully open-source systems using state-of-the-art methods. In our work, we construct an extensible and robust spoken dialogue system as a conversation partner. 
Figure~\ref{fig:changes} shows an overview of EDEN, which makes several key improvements over an empathetic chatbot architecture proposed by \citet{siyan2024using}.\footnote{We further discuss the original chatbot design in Appendix \ref{app:chat_details} and more information can be found at the original paper's site \url{https://github.com/siyan-sylvia-li/adaptive_empathetic_BEA2024}} 
For each turn, the user input is first analyzed for negative sentiment and prolonged pauses. If these signals are captured, corresponding empathetic feedback is synthesized. Otherwise, a grammatical feedback module utilizes a grammar correction model to locate grammar errors and construct feedback using templates. We further devise a grammar correction hierarchy to not overwhelm users (Appendix \ref{app:grammar_correction}). When the user utterance does not trigger grammatical feedback, the conversation proceeds as normal through a conversation language model. 

 % partially adopt chatbot architecture from \citet{siyan2024using} with improvements in multiple areas for our experiments (Figure \ref{fig:changes}).

After the user receives feedback, they occasionally have follow-up queries. We resort to ChatGPT for resolving these queries if they are relevant to English learning. We use ChatGPT for this purpose rather than our conversation model because responding to user queries about the feedback would be out of the scope of our conversation model. Instead, we employ a transition module (Appendix \ref{app:transitions}) to continue the original conversation after ChatGPT responses. Additional design choices are informed by our chatbot design survey conducted on Twitter / X with more than 450 responses (Appendix~\ref{app:discussion}). 

We make several innovations in our system design to strengthen EDEN for \textit{spoken}, \textit{educational}, and \textit{open-domain} English-practice dialogue. We discuss how we create a grammar feedback model for spoken utterances (Section \ref{sec:grammar_tailor}) and an open-domain chit-chat model (Section \ref{sec:convo_model}), detail the construction of adaptive empathetic feedback (Section \ref{sec:adaptive_emp}), and specify how users can customize EDEN according to their needs (Section \ref{sec:personalization}).
% Although the design was interesting, ... we strengthen the previous design...
% Despite positive results during an initial user study, the chatbot design can be further improved in the following areas:
%
% What's the entire work flow like, directly do workflow, directly use the chart, highlight the differences, remove paragraphs about the things to improve
%
% Smaller 
%
% \noindent \textbf{Better grammatical feedback -} 
%
% \noindent \textbf{Choice over conversation topic -} 
%
% \noindent \textbf{More personalized empathetic responses -} The authors separately examine whether GPT-4 judge preferences align with human preferences for different versions of generated empathetic responses. The results suggest no significant alignment between user and GPT-4 preferences. In particular, some users simply prefer shorter responses, while others find the praises in the feedback disingenuous. Overall, users exhibit a variety of preferences for empathetic feedback\footnote{We have contacted the authors and discussed results that were not mentioned in the original paper}. Appropriate personalization should be able to address this issue.
%

% to contribute to NLP in education.  

\subsection{Tailored Grammatical Feedback for Spoken Utterances}
\label{sec:grammar_tailor}

Prior work on grammar corrections focuses on written texts rather than dialogue \cite{yasunaga2021lm,bryant-briscoe-2018-language,katinskaia-yangarber-2023-grammatical}. 
\citet{yuan-etal-2022-eracond} presents a model trained on dialogue utterances, but the collected dataset focuses on written dialogue, which can differ from spoken dialogue. 
There is a lack of dialogue grammar correction datasets for spoken conversations. We bridge this gap by training a grammar correction model specialized for transcribed spoken utterances. Since the grammar correction process occurs for every user utterance, and constant GPT-4 API queries are costly and inefficient, we use a local model instead of GPT-4. 

\subsubsection{Grammar Correction Data}

To train such models, we utilize a set of Whisper-Medium-transcribed audio clips collected from an existing English learning chatbot platform \cite{li2022using} proposed by \citet{siyan2024using}. The speakers of these clips are native Mandarin speakers, which is representative of our target users. We adopt this set for training a grammar correction model specialized for spoken utterances.

Since there is a lack of ground-truth grammar corrections for these utterances, and language models such as GPT-4 are effective in grammar error correction tasks \cite{kobayashi2024large,coyne2023analyzing}, we use GPT-4-Turbo to generate grammar corrections for these transcriptions. 
% During this process, GPT-4-Turbo often assesses the grammatical correctness of the grammatically correct sentences instead of correcting (i.e. responding with "the sentence X is grammatically correct"). We adjust our pipeline by using the original sentence as the correction when GPT-4-Turbo deems this sentence grammatically correct. 
% For utterances with multiple sentences, we first split them into individual sentences using the \texttt{nltk} toolkit to be consistent with test-time usage. 
In total, we produce 7,894 sentence-correction pairs.

\begin{table*}[!h]
    \centering
    \begin{tabularx}{\textwidth}{bbb}
        \toprule
        \textbf{Original}     & \textbf{Flan-T5-XL}     & \textbf{Llama-2}     \\
        \midrule
        % Like during the lunch break, she would actually brought me back to her office and she would share with me a lot of snacks she loved and she also told me whenever I felt lonely, I could talk with her.         & During the lunch break, she would actually bring me back to her office and share with me a lot of snacks she loved. She also told me that whenever I felt lonely, I could talk with her.        & During the lunch break, she actually brought me back to her office, where she shared a lot of snacks with me that she loved. She also told me that whenever I felt lonely, I could talk with her.       \\
        % \midrule
        It's not famous, but I really love it because it's my hometown, and I live here until now with my family and friends.         & It's not famous, but I really love it because it's my hometown, and I've lived here until now with my family and friends.
        & It's not famous, but I really love it because it's my hometown and I still live here with my family and friends.         \\
        \midrule
        Dry weather will give a dry skin and some illness comes from it. &	Dry weather causes dry skin, and some illnesses result from it.	& Dry weather will give dry skin and lead to some illnesses. \\
        \bottomrule
    \end{tabularx}
    \caption{Example grammar corrections generated by our two models.}
    \label{tab:grammar_sample}
\end{table*}

\begin{table}[!t]
    \centering
    \begin{tabular}{ccc}
    \toprule
         & \textbf{\#L-Correct} & \textbf{\#L-Wrong}\\
        \midrule
       \textbf{\#F-Correct}  & 61.6\% & 11.1\%\\
       \textbf{\#F-Wrong} & 12.9\% & 14.4\%\\
    \bottomrule
    \end{tabular}
    \caption{Percentages of validation set transcribed utterances that \textbf{Llama-2} generates valid and invalid corrections for (\#\textbf{L}-Correct and \#\textbf{L}-Wrong), as well as the number of utterances that \textbf{Flan-T5-XL} provides valid and invalid corrections for (\#\textbf{F}-Correct and \#\textbf{F}-Wrong), as measured by \textbf{GPT-4}.}
    \label{tab:gpt_grammar_numbers}
\end{table}

\begin{table}[]
    \centering
    \begin{tabular}{cccc}
    \toprule
        \textbf{Study} & \textbf{Win\%} & \textbf{Lose\%} & \textbf{Tie\%} \\
    \midrule
    Study \#1, \textit{All} & \textbf{46.3} & 35.8 & 17.9 \\
    Study \#1, \textit{Maj} & \textbf{45.0} & 34.5 & 20.5 \\
    \midrule
    Study \#2, \textit{All} & 39.3 & \textbf{40.3} & 20.4 \\
    Study \#2, \textit{Maj} & 40.1 & \textbf{41.4} & 18.5 \\
    \bottomrule
    \end{tabular}
    \caption{\textbf{Llama-2 vs. Flan-T5} win, lose, and tie rates from the two \textbf{human subject} studies. \textit{All} indicates including all sentences, and \textit{Maj} indicates the results when only including sentences with a preferred majority (more than 50\% of participants voted for one option) among participants.}
    \label{tab:grammar_results}
\end{table}

\subsubsection{Model Training and Evaluation}
We train two models, a Llama-2 7B model and a Flan-T5-XL model, on the same data. Both models are trained on a single GPU using Parameter-Efficient Fine-Tuning \cite{peft}. See further training details in Appendix \ref{app:grammar_model_training}.

Due to a lack of ground-truth data, we again use GPT-4-Turbo to compare model performance on the validation set. For sentences where the two models disagree, we prompt GPT-4-Turbo to assess whether the model corrections are grammatically correct. We report the judgment results on these utterances in Table \ref{tab:gpt_grammar_numbers}. We find that the two models usually both provide valid corrections, with the Llama-2 model slightly out-performing.
% Out of 810 validation set entries, the two models make different corrections on 380. For the other 430 sentences where both models make the same correction, 389 are correct according to GPT-4.

For a more robust evaluation, we conduct two human-subject studies by recruiting participants from \href{https://www.prolific.com/}{Prolific}. Our goal is to compare the two models \textit{when they differ}, under two conditions: (i) when at least one reports a grammatical error, which essentially assesses their \textit{precision} (Study \#1), and (ii) when an expert identifies a grammatical mistake in the original utterance, which essentially assesses their \textit{recall} (Study \#2). For Study \#1, a random subset of 40 medium-length sentences from the validation set. For Study \#2, we curate 31 such sentences that are grammatically incorrect. In both studies, six native English speaker participants, paid at \$15 per hour, compare the corrections for each sentence from the two models. The participants are asked to select the better correction, defined as (1) minimally changing the original sentence, (2) retaining the original meaning, and (3) grammatically correct. Each participant evaluated 20 sentences for Study \#1 and 31 sentences for Study \#2.

Since both models tend to generate valid corrections, participant preferences vary, resulting in low inter-annotator agreement. For the first study, the Fleiss' kappas for two 20-sentence batches are 0.310 and 0.301 respectively (fair agreement). For the second study, the kappa is 0.139 (limited agreement). Therefore, we additionally evaluate participant preferences for sentences with a majority of participants agreeing (Table~\ref{tab:grammar_results}). 

While participants prefer Llama-2 for randomly selected transcripts, they slightly prefer Flan-T5-XL for erroneous sentences. This could be due to Llama-2 providing higher-quality rewrites to grammatically correct sentences. Considering that Llama-2 is generally preferred in Study \#1 and the differences in Study \#2 are minor, we choose the fine-tuned Llama-2 as the grammar model.

\subsection{Open-Domain Conversation Model}
\label{sec:convo_model}
% \Sylvia{Updated training data, what is the topic selection process, make sure to include training details, including issues we found with the original prompting approach, analyses of Chinese students versus ESL, e.g. Chinese songs only}
Previous English chatbots designed for speaking practice focus on delivering course content in formal English training \cite{DU2024100230}. Here, we are targeting users who are learning English out of interest. Therefore, EDEN must engage users in interesting conversations to improve their experience and reduce their language anxiety \cite{von2003students}. 
% As a result, an open-domain chitchat model better suits our needs. 
We create a conversation model capable of discussing various topics to accommodate user interest. Although ChatGPT or GPT-4o is a convenient choice, GPT-4o can be too slow for chit-chat and is not widely accessible in China, where some of our recruited users are located. 
% ChatGPT often produces verbose responses that are unnatural in everyday conversation. 
% We choose not to use ChatGPT directly for conversations due to cost and efficiency concerns.

We adapt a data synthesis pipeline with persona-based prompting \cite{li2023curriculum} to support social chit-chat while retaining its strength in generating naturalistic and accessible conversation responses and taking initiatives. 
% Additionally, since generated personas are incorporated into the training data synthesis pipeline, the resulting training data is more diverse, increasing the robustness of the fine-tuned model. Based on these advantages, we decided to adapt the data synthesis pipeline from the original paper. 
Specifically, we use everyday topics (e.g. favorite cuisine), remove some extraneous constraints in the original pipeline, and adjust prompts after analyzing preliminary outputs.

% \noindent \textbf{Topic Selection:} 
The following broad topic areas are used: \textit{food, books, movies, TV shows, music, hobbies, and English learning}. 
% This wide coverage ensures that users can converse about their topics of interest. 
We further identify 243 relevant topics within these broad areas (Appendix~\ref{app:convo_topics}).

We ask ChatGPT to generate 10 two-party conversations per topic. One of the personas is generic (often assumed to be American by ChatGPT), and the other is someone whose first language is not English. We originally used a hypothetical Chinese college student as the second persona to be consistent with prior work, but the generation diversity was problematically low (Appendices \ref{app:gen_div_issues} and \ref{app:synth_prompts}).
% Prior research indicates that discussing topic of interest is correlated with lower foreign language anxiety \Sylvia{Anxiety literature}, so

Upon further examination of the generated data, we discover some low-quality generated conversations, which we address through filtering. These lower-quality conversations often feature one conversation party making assumptions about the other party. See details about these quality issues in generated conversations, as well as the data filtering process, in Appendix \ref{app:data_filter}.
% We incorporate a filtering mechanism to prevent the fine-tuned model from making similar mistakes. Given a conversation, we ask ChatGPT whether Person 1 makes assumptions about Person 2 or fails to make direct, specific recommendations when requested. If so, then the conversation is pruned.
After filtering, 1,227 conversations remain. The conversations have an average length of 8.35 turns. 
% Similar to \citet{li2023curriculum}, we reformat the conversation such that Person 1 is now ASSISTANT and Person 2 is now USER. We additionally prepend a USER initial turn to each conversation that specifies the current ASSISTANT persona and topic.
We then fine-tune a Llama-2 model on this conversation data (See Appendix~\ref{app:conv_model_training} for details). Please see the distribution of topics among the conversations in Appendix~\ref{app:topic_dist}.

% \subsubsection{Inference Time Usage}

% During our experiments, participants choose between the broad topic areas. Each topic area corresponds to a randomly selected relevant topic. We fix the persona to be an Asian-American female.
% To prevent the chatbot from forgetting the conversation history due to a finite context window, we use ChatGPT to summarize the current conversation every seven turns. 

\subsection{Adaptive Empathetic Feedback}
\label{sec:adaptive_emp}

Following \citet{siyan2024using}, EDEN's empathetic feedback mechanism triggers when the system registers signals of user distress such as heightened negative affect or prolonged pauses. A ChatGPT prompt, optimized through the DSPy framework \cite{khattab2023dspy}, is used to produce a piece of feedback from past user utterances. Generally, the feedback (i) sounds empathetic and colloquial, (ii) includes examples and actionable feedback. Since the generated feedback can still sound overly formal, we use additional rewrite prompts to shorten the feedback and reduce its formality. 
% Personalization, if applicable, is then applied to this tailored feedback. 

\subsection{Personalization Feature}
\label{sec:personalization}
We notice in our design survey that users have a variety of preferences. Two design aspects that reflect such diversity are whether to include Mandarin translations of chatbot utterances and the length of chatbot feedback. We thus allow users to customize EDEN by including personalization questions in our experiment flow before any conversations:

\noindent \textbf{Q1:} Do you want Mandarin translations of the chatbot utterances? (Yes / No)

If the participant selects Yes, each chatbot utterance is translated into Mandarin using ChatGPT.

\noindent \textbf{Q2:} How would you like the chatbot feedback? (Succinct / Details \& examples / No preference)

Adaptive empathetic feedback utterances are customized through prompting using user responses. The original generated feedback is used if the participant has no preference (Appendix~\ref{app:personalization}).

\section{User Study: Empathetic Feedback}
We recruited 31 native Mandarin speakers from the internet (15) and the authors' home institution (16). The internet participants were not compensated, while the participants from the author's institution received \$15 Amazon gift cards. Our IRB-approved study intends to answer these research questions: 

\noindent \textbf{RQ1:} Does adaptive, empathetic feedback in an English-teaching chatbot result in higher perceived affective support (PAS)?

\noindent \textbf{RQ2:} Does higher chatbot PAS correlate to positive changes in L2 grit?

\subsection{Experimental Conditions}
Our participants are assigned to one of three experimental conditions sequentially:
\begin{enumerate}
    \item \textit{No Empathetic Feedback (\textbf{None})}.
    \item \textit{\textbf{Fixed} Empathetic Feedback:} The empathetic feedback is randomly selected from a pre-defined list of generic empathetic phrases. See the fixed empathetic responses in Appendix~\ref{app:fixed_emp}.
    \item \textit{\textbf{Adaptive} Empathetic Feedback:} The empathetic feedback is generated through ChatGPT using prompts from \citet{siyan2024using}. We personalize this feedback.
\end{enumerate}

All other components of EDEN (i.e. grammatical feedback, conversation, etc.) are held constant across conditions. By defining the conditions as such, we can more rigorously test the effect of the presence and the different types of empathetic feedback on student L2 grit and chatbot PAS. We hypothesize that: 

\noindent \textbf{H1:} Both Fixed and Adaptive conditions improve chatbot PAS, but Adaptive is more effective.

\noindent \textbf{H2:} Higher chatbot PAS would correlate to positive changes in L2 grit. 

\subsection{Experimental Procedure}
The participants first complete a pre-survey about their English proficiency and L2 grit \cite{teimouri2022l2}. They then proceed to converse with EDEN for at least three conversations after completing the short personalization questionnaire. Upon completion of the chatbot interaction phase, the participants evaluate their experience, the chatbot's PAS, and their L2 grit in a post-survey. We use the same adapted chatbot PAS survey \cite{siyan2024using} and the L2 grit survey \cite{teimouri2022l2}. All questions are five-item Likert-scale questions presented in both English and Mandarin.

\begin{table}[!h]
    \centering
    \begin{tabularx}{\linewidth}{b}
    \toprule
       \textbf{Question Code: Question Text} \\
    \midrule
       \textbf{QUAL:} How was the conversation quality? \\
        \textbf{CONF:} Do you feel more confident after conversing with the chatbot? \\
        \textbf{USE:} Do you think the chatbot's grammar feedback is useful? \\
       \midrule
       \textbf{ENC:} The chatbot encourages me when I am having difficulties in the conversation. \\
       \textbf{LIST:} The chatbot listens to me when I have something to say. \\
       \textbf{CARE:} My opinion matters to the chatbot. \\
       \textbf{APP:} The chatbot recognizes and appreciates when I am good at something. \\
       \midrule
       \textbf{L2.1:} I am a diligent English language learner.\\
       \textbf{L2.2:} My interests in learning English change from year to year. \\
       \textbf{L2.3:} When it comes to English, I am a hard-working learner. \\
       \textbf{L2.4:} I think I have lost my interest in learning English. \\
       \textbf{L2.5:} Now that I have decided to learn English, nothing can prevent me from reaching this goal. \\
       \textbf{L2.6:} I will not allow anything to stop me from my progress in learning English. \\
       \textbf{L2.7:} I am not as interested in learning English as I used to be. \\
       \textbf{L2.8:} I was obsessed with learning English in the past but have lost interest recently. \\
       \textbf{L2.9:} I put much time and effort into improving my English language weaknesses. \\
    \bottomrule
    \end{tabularx}
    \caption{Select questions used for measuring general conversation quality (top), PAS (middle), and L2 grit (bottom) in the pre- and post-surveys. We provide additional survey details in Appendix~\ref{app:pre_post_survey}.}
    \label{tab:questions}
\end{table}

\begin{figure*}[!ht]
    \centering
    \includegraphics[scale=0.65]{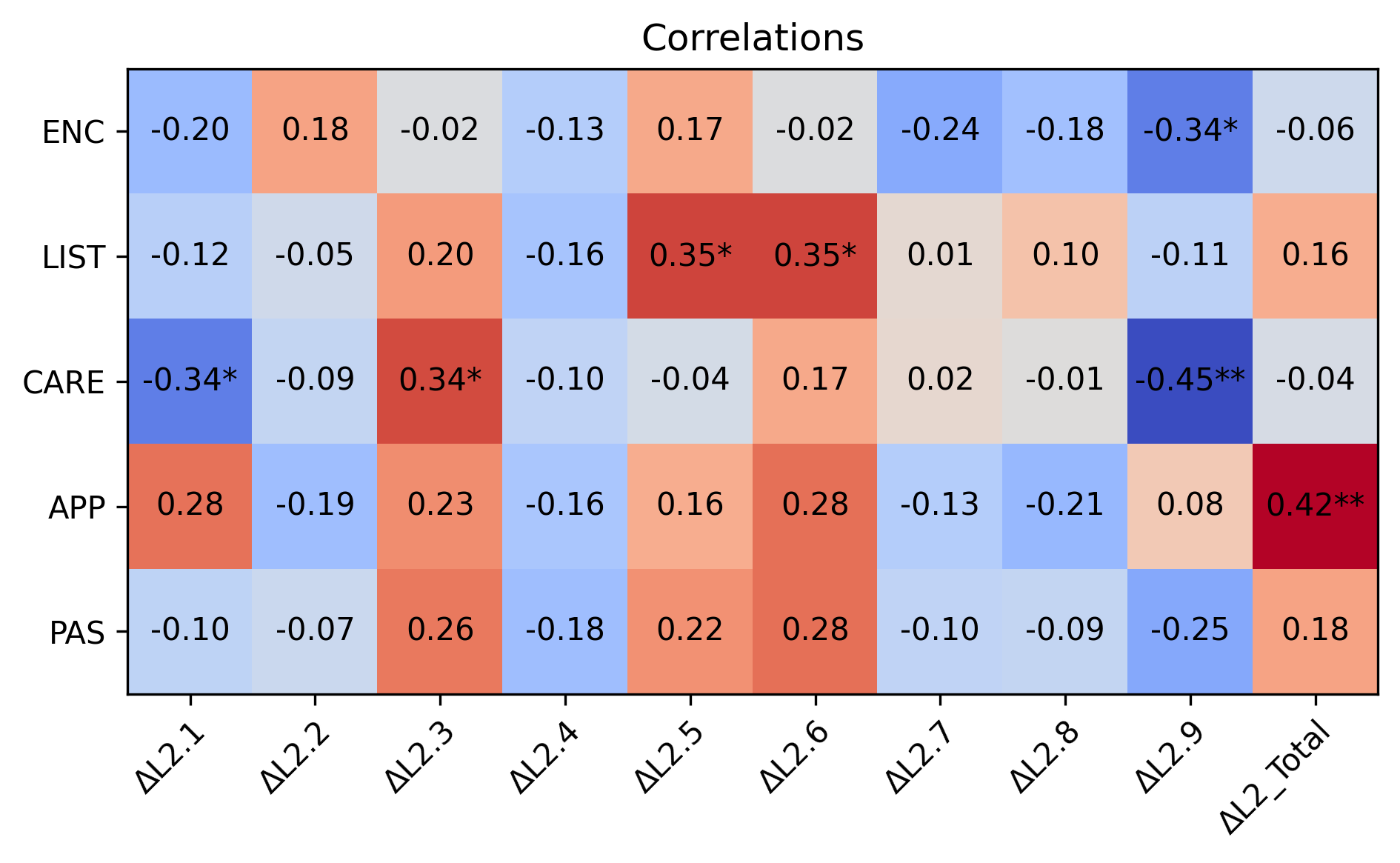}
    \caption{Correlations between different measures for chatbot PAS and L2 grit changes. **$ = p < 0.05$, *$ = p < 0.1$. }
    \label{fig:corr_matrix}
\end{figure*}
\begin{table*}[!ht]
    \centering
    \begin{tabular}{lcccccccccc}
    \toprule
    $\Delta$ for & \textbf{L2.1}  & \textbf{L2.2} & \textbf{L2.3} & \textbf{L2.4} & \textbf{L2.5} & \textbf{L2.6} & \textbf{L2.7} & \textbf{L2.8} & \textbf{L2.9} & \textbf{L2$_{Total}$}\\
     \midrule
     \textbf{None} &0.05 &-0.24 &-0.05 &0.29 & -0.12&-0.24 &0.24 &0.47 &0.47 &-0.64\\
     \textbf{Fixed} & \textbf{0.83} &0.67 &\textbf{0.33} &-0.17 &\textbf{0.17} &0.00 &\textbf{-0.17} &-0.33 &\textbf{0.83} &\textbf{2.17}\\
     \textbf{Adaptive} &0.25 &\textbf{-0.50} &0.13 &\textbf{-0.25} &0.13 &\textbf{0.25} &-0.13 &\textbf{-0.38} &0.13 &2.13\\
     \bottomrule
    \end{tabular}
    \caption{Average L2 grit changes for the different experimental conditions. Note that items 2, 4, 7, and 8 are reverse-coded, so lower would be better for these items.}
    \label{tab:result_l2}
\end{table*}

\begin{table}[]
    \centering
    \begin{tabular}{lccccc}
    \toprule
     & \textbf{ENC} & \textbf{LIST} & \textbf{CARE} & \textbf{APP} & \textbf{PAS}\\
     \midrule
     \textbf{None} &3.53 &\textbf{4.12} &\textbf{4.00} &3.47 & 3.78 \\
      \textbf{Fixed} &3.83 &2.83 &3.00 &3.67 &3.33 \\
     \textbf{Adap.} & \textbf{4.38} &4.00 &3.88 &\textbf{4.38} & \textbf{4.16} \\

     \bottomrule
    \end{tabular}
    \caption{Average PAS questionnaire results for the different experimental conditions.}
    \label{tab:result_pas}
\end{table}

Overloading the question codes to be the reported values for their questions, we further define:
\begin{align*}
    \textit{PAS} &= \frac{\textit{ENC} + \textit{LIST} + \textit{CARE} + \textit{APP}}{4} \\
    \Delta \textit{L2.k} &= \textit{L2.k}_{post} - \textit{L2.k}_{pre}
\end{align*}

where $\textit{L2.k}_{post}$ and $\textit{L2.k}_{pre}$ refer to the reported values for the L2-grit-related question \textit{L2.k} in the post-survey and the pre-survey, respectively. Since the survey has items 2, 4, 7, and 8 reverse-coded, we compute the total change in L2 grit as such:
\begin{align*}
    \Delta \textit{L2}_{Total} &= \Delta \textit{L2.1} - \Delta \textit{L2.2} + \Delta \textit{L2.3} \\
    & - \Delta \textit{L2.4} + \Delta \textit{L2.5} + \Delta \textit{L2.6} \\
    & - \Delta \textit{L2.7} - \Delta \textit{L2.8} + \Delta \textit{L2.9}
\end{align*}

\section{Results and Discussion}

Our participants display intermediate self-reported English proficiency. They have studied English for an average of 15.9 years. Since Chinese citizens tend to start learning English at a young age, this number is not out of the ordinary. Their average scores for IELTS and TOEFL are 6.7 and 110.6, respectively. Most participants speak more Mandarin than English in their everyday lives. 

On average, the participants conversed with our chatbot for 31.19 turns\footnote{We were not able to locate two None condition participants' conversation data due to experiment ID mismatch. }. They received 1.57 grammatical feedback during the interactions, and participants under Fixed and Adaptive conditions received 4.42 and 2.67 empathetic feedback, respectively. The top three selected topics are food, English learning, and books. 

Due to our participants having intermediate English proficiency on average, some participants do not trigger the empathetic feedback module. We therefore reassign participants who did not trigger empathetic feedback to None condition. After this reassignment, we have 17 participants for the None condition, six for Fixed, and eight for Adaptive. 

Overall, the participants consider the conversations to be moderate-to-high quality ($\overline{\text{QUAL}} = 3.39$). They experience some confidence boost post-interaction ($\overline{\text{CONF}}=3.39$), and they find the grammar feedback useful ($\overline{\text{USE}} = 3.52$). Some conversation quality ratings were negatively affected by network errors during experiments.

\subsection{Causal Relationship between Empathetic Feedback and PAS}
%JH: casual or causal?
%SL: causal, oops
We present the post-survey results for chatbot PAS for the different conditions in Table~\ref{tab:result_pas}. Adaptive outperforms Fixed for all PAS-related metrics. This is expected, as a generic phrase is unlikely to elicit as much perceived empathy as a tailored, adaptive piece of feedback. Furthermore, using fixed phrases may be perceived as more unnatural in a conversation than using a personalized response, making the participants feel not listened to. The Adaptive condition results in the highest PAS in the pilot study, although this dominance does not persist across different items. Users feel more listened to and that their opinions matter more under the None condition potentially for a similar reason; the current potentially unnatural transition between dialogue and feedback content can lead to the users perceiving a lack of chatbot attention. Adaptive performs the best in encouragement and appreciation, and both Adaptive and Fixed are better than None here. This could indicate that EDEN's empathetic feedback mechanism correctly identifies and addresses participant struggles. Including praises in the empathetic feedback pipeline likely contributes to a higher appreciation rating. Our adaptive condition also causes a higher PAS rating compared to what \citet{siyan2024using} reported, which is 3.27, highlighting our improvement.

These results validate the first hypothesis. We postulate that by making EDEN's transition between conversation and feedback more seamless, we can enhance perceived affective support further by helping users feel better attended to.

\subsection{Correlation between PAS and L2 Grit}
Table \ref{tab:result_l2} records the average L2 grit changes per condition. We notice that the None condition never achieves the most positive L2 grit changes. Meanwhile, although the Fixed condition is associated with the lowest PAS, it achieves the highest overall L2 grit changes, slightly above Adaptive.

Pearson's correlation is used to evaluate the relationship between various components of PAS and changes in L2 grit (Figure~\ref{fig:corr_matrix}). Our results showcase some components of PAS being weak to intermediate predictors for positive L2 grit changes. Specifically, perceived chatbot appreciation correlates positively with changes in total L2 grit, and users feeling their opinions matter predicts positive changes in self-perception of being hard-working. Additionally, users feeling that they are listened to is correlated with increased self-determination.

We identify counter-intuitive results that can be attributed to our small sample size. By the L2 grit questionnaire definition, $\Delta$L2.1 and $\Delta$L2.3 should be positively correlated. However, users feeling their opinions matter positively correlates with one and not the other. Similarly, it negatively correlates with $\Delta$L2.9, the self-perception of putting much effort into improving English skills. Higher PAS still weakly correlates with positive L2 grit changes, suggesting that our results partially align with \citet{wu2023teachers} and supporting our second hypothesis.

\subsection{Additional Correlations}
\textbf{PAS and Conversation Quality:} During the user study, we noticed that negative bot interactions can reduce PAS. We are therefore curious about how PAS relates to conversation quality ratings. We find significant positive correlations between PAS and the conversation quality measures (Appendix~\ref{app:pas_conv}). This result suggests using PAS as a reliable conversation quality measure for social chatbots. 

\noindent \textbf{English Proficiency and L2 Grit Changes:} We discover that English proficiency does not significantly correlate with a total of L2 grit changes. This indicates that, in our pilot study, being more proficient does not preclude users from having higher L2 grit after chatbot interactions. If this result generalizes, English learners from all levels could benefit from chatbot interventions that improve grit. 

\noindent \textbf{PAS and L2 Grit:} Directly reproducing \citet{wu2023teachers} results, we examine the correlation between various PAS measures and \textit{L2 grit} in the post-survey (Appendix~\ref{app:pas_grit}). We find that higher perceived appreciation is an intermediate-strength predictor for higher L2 grit in the post-survey. This result partially validates the generalizability of \citet{wu2023teachers} results to chatbot settings, since only perceived appreciation serves as a sufficient predictor, and overall PAS has no significant correlations with any of the L2 grit measures.

\subsection{User Feedback}

Users generally appreciate the quality of recommendations made by EDEN and the naturalness of chatbot responses. One user commented that EDEN provides recommendations highly tailored to their preferences. Several users commended how engaging the conversations were. Another user acknowledged the benefits of using a chatbot as a language practice partner: \textit{"Notably, I felt more at ease communicating with the AI than with a human, as there is often a fear of judgment regarding one's speaking abilities."} A few users indicated excitement about trying EDEN in the future as a commercial product. However, some participants dislike the grammar feedback since they already have high English proficiency and do not require the more basic feedback. 

Participants assigned to both empathetic conditions enjoyed the chatbot's supportiveness. One of the participants under the adaptive condition said, "\textit{I was rather surprised when I received the encouraging feedback but in a good way.}" 

Currently, EDEN has little capability beyond social chitchat and providing grammatical and empathetic responses. Therefore, it would fail when users request their English skills to be evaluated (e.g. "How good do you think my English speaking skill is?" or "Could you evaluate my English skill") or query the number of turns in the current conversation. These requests could signal participants placing trust in EDEN's capabilities. Another failure mode occurs when the chatbot uses vocabulary beyond the users' comprehension. Future work can address this by developing additional functionalities and user-adaptive mechanisms for vocabulary choice. We provide example conversations in Appendix~\ref{app:example_convo}.

\section{Conclusion}

In this work, we build EDEN, a robust open-domain empathetic English-teaching chatbot tailored for spoken conversations. We then use it to verify whether results from \citet{wu2023teachers} extend to perceived affective support of chatbots in addition to teachers. Our initial user study reveals that higher perceived affective support of our chatbot correlates positively with changes in student L2 grit. We additionally showcase that adaptive, empathetic feedback surpasses fixed and no empathetic feedback in enhancing chatbot perceived affective support. Our work serves as a first step in exploring dialogue system interventions for boosting L2 grit.

% For future work, we intend to demonstrate the effectiveness of the adaptive empathetic framework across different languages for language education. We have also planned a larger-scale user study to further and better validate our results. 

\section{Limitations}
Our human evaluation has several limitations. Due to a lack of convenience samples, we had to recruit from the internet and the authors' home institution which is a university in the United States. This poses a sampling bias since individuals recruited through these channels tend to have at least moderate English proficiency; thus the diversity in our sample is inherently limited. Furthermore, since we were not compensating our online user study participants, it was difficult to retain these participants. Another issue with our experimental results lies in the imbalance of participants in the different experimental conditions. Since our empathetic intervention is based on participants exhibiting distress or prolonged pauses in speech, it is inherently more frequently triggered with more beginner-level participants, which does not align with the typical Mandarin native-speaker population residing in the US. We have attempted to expand our experiment to Mainland Chinese users, but they tend to experience connection issues to our servers. We currently only target native Mandarin speakers because the system from \citet{siyan2024using} is developed from Mandarin speaker data and focuses on native Mandarin speakers. We may experiment with expanding to all English-as-a-second-language individuals in the future. 

In terms of implementation, areas such as Text-to-Speech synthesis could use further improvement, as the perception of empathy often hinges upon the tone of the voice. Furthermore, as discussed in the user feedback section, some users have asked for assessments of English skills and the definition of the assessments, as well as how long the conversation has been, which we do not have a standardized set of guidelines for. Future work could incorporate further modularization with different functionalities. There were also intermittent server connection issues on the users' ends which negatively impacted their experience.

\section{Ethical Considerations}
As we observe in the study, some users would assign authority to the chatbot and request feedback. When the feedback is not truthful or not sufficiently encouraging, the users may experience negative emotions as a result. We must additionally ensure minimizing and eliminating any harmful content the chatbot could produce when interacting with users. Users anthropomorphizing our chatbot is beneficial for perceiving empathy from the chatbot, but we should strike a delicate balance to avoid users becoming emotionally dependent or attached.

Using automated systems for education purposes might raise concerns about these pedagogical systems replacing teachers. We would like to state that our chatbot is intended to be a conversation practice partner outside of the classroom, and is not a replacement for human instruction.

\bibliography{anthology,custom}

\appendix

\section{Chatbot Design from \citet{siyan2024using}: More Details}
\label{app:chat_details}
\citet{siyan2024using} proposes a novel adaptive and empathetic English-teaching chatbot. The chatbot detects heightened negative emotions and prolonged pauses in student speech using an existing wav2vec 2.0 speech emotion detection model \cite{baevski2020wav2vec} and a voice activity detection toolkit \cite{Silero-VAD}. When these signals are captured, ChatGPT is used to generate empathetic feedback using the past three student utterances. The authors use DSPy \cite{khattab2023dspy} to optimize their ChatGPT prompt such that the resulting colloquial feedback sounds empathetic and contains specific examples. 

In addition to providing empathetic feedback, the chatbot offers grammatical feedback on student utterances using a fine-tuned Llama-2 \cite{touvron2023llama} model for grammar correction. The grammar correction model is trained on the ErAConD dataset \cite{yuan-etal-2022-eracond}, which contains written dialogue utterances and their expert grammar corrections. The SERRANT \cite{choshen2021serrant} package is then used to locate the specific grammatical errors by comparing the generated correction and the original sentence. The grammatical feedback is a combination of a rephrase (e.g. \textit{Maybe you meant "had" rather than "has"}) and a template-based explanation for each error type \cite{liang-etal-2023-chatback}.

A transition module is designed to aid smoother transitions between different chatbot system components. It connects empathetic or grammatical feedback to the original conversation. The module classifies whether student utterances constitute an English-learning-related query using pre-written rules. If a student utterance is classified as a relevant query, the system prompts ChatGPT to generate an answer for the query, otherwise, a randomly selected pre-defined connector phrase is used to transition back to the original conversation directly. 

\section{Chatbot Design Survey}
\label{app:discussion}
We polled 456 Mandarin-speaking users on Twitter / X for their opinions on designing an empathetic English-teaching chatbot. The survey contains items regarding features not included in \citet{siyan2024using}, such as Mandarin translations for chatbot utterances, as well as items similar to the original survey.

We present the questions and the responses from the survey below.

\begin{enumerate}
    \item How do you like the tone of your English teacher's feedback to be?
    \begin{enumerate}
        \item \textbf{Colloquial (80\%)}
        \item Formal (20\%)
    \end{enumerate}
    \item How long should teacher feedback be?
    \begin{enumerate}
        \item 1 - 2 sentences (21.3\%)
        \item \textbf{2 - 3 sentences (52.6\%)}
        \item 3 - 4 sentences (13.2\%)
        \item 4+ sentences (12.9\%)
    \end{enumerate}
    \item If you made a mistake, how would you like your errors to be corrected? Select all that apply.
    \begin{enumerate}
        \item Correct your errors directly (38.8\%)
        \item Help you self-correct your errors using questions (43.6\%)
        \item \textbf{Give you examples such that you can learn from these examples and avoid making the same errors again (80.5\%)}
    \end{enumerate}
    \item What does an ideal encouraging feedback from English teachers look like? Select all that apply.
    \begin{enumerate}
        \item Give you encouragement, such as "You are doing great!" or "I am proud of you!" (38.2\%)
        \item Tell you what you are good at in spoken English (41.9\%)
        \textbf{\item Tell you what you can do to improve your spoken English (72.4\%)}
        \item \textbf{Tell you how to improve your spoken English through examples (78.7\%)}
        \item Give you practical advice for English learning (43.4\%)
    \end{enumerate}
    \item Our current chatbot design contains a button that reveals the transcript of the chatbot utterance when clicked; should we keep this button?
    \begin{enumerate}
        \item \textbf{Yes! (87.9\%)}
        \item No, the transcript should be displayed directly and automatically (12.1\%)
    \end{enumerate}
    \item Do you need Mandarin translations of chatbot utterances?
    \begin{enumerate}
        \item \textbf{I only need translations for chatbot feedback (38.2\%)}
        \item I need translations for everything that the chatbot says (28.5\%)
        \item I don't need any translation (33.3\%)
    \end{enumerate}
\end{enumerate}

There are some additional free-form responses provided by the internet users filling out our survey. We intend to perform further analyses of the survey and publicly share the results to provide research directions for others in the field. 

\section{Grammar Model Training Details}
\label{app:grammar_model_training}
We use a train-validation split of 0.9-0.1 when training our models. Both models were fine-tuned using PEFT on a single GPU for 10 epochs. The Llama-2 7B model was trained with an initial learning rate of \texttt{2e-4} and a batch size of 4. The Flan-T5-XL grammar model was trained with default parameters. The best checkpoints according to evaluation losses were selected. 

\section{Conversation Model Training Details}
\label{app:conv_model_training}
For the conversation data, we use a train-validation split of 0.95-0.05. The Llama-2 model was PEFT-trained on a single GPU for 10 epochs, with an initial learning rate of \texttt{2e-4} and a batch size of 4. The best checkpoint according to evaluation losses was selected. 

\section{Grammar Correction Hierarchy}
\label{app:grammar_correction}

We reference an online resource for grammatical error hierarchy\footnote{\url{https://bcourses.berkeley.edu/courses/1196299/files/66663155/download?wrap=1}} to establish the hierarchy of grammar errors recognized by our system.

In this hierarchy, errors are divided into tiers based on severity. Different error tiers correspond to different tolerance levels. For instance, if an error is tier \#1 with a tolerance level 1, the error is immediately corrected (grammatical feedback is given on this error); if an error is tier \#3 with a tolerance level 5, then this error will only be corrected if the user has made the same type of error for five times in one conversation. We detail this hierarchy in Table~\ref{tab:grammar_hierarchy}.

\begin{table}[!h]
    \centering
    \begin{tabularx}{\linewidth}{sbs}
    \toprule
        \textbf{Tier} & \textbf{Errors} & \textbf{Tol.} \\
    \midrule
    \#1 & Word Order, Wrong Verb Tense, Incorrect Verb Form, Incorrect Preposition, Missing Preposition, Unnecessary Preposition, Wrong Collocation & 1 \\
    \midrule
    \#2 & Subject-Verb Disagreement, Incorrect Singular/Plural Noun Agreement, Incorrect Possessive Noun, Incorrect Determiner& 3 \\
    \midrule
    \#3 & Incorrect Auxiliary Verb, Incorrect Part of Speech, Missing Word Related To Verb Form, Missing Word Related To Verb Tense, Missing Determiner, Missing Verb, Missing Adjective, Missing Adverb, Missing Auxiliary Verb, Missing Adpositional Phrase, Missing Conjunction, Missing Particle, Missing Noun, Missing Pronoun, Unnecessary Determiner, Unnecessary Verb, Unnecessary Word Related To Verb Form, Unnecessary Word Related To Verb Tense, Unnecessary Adpositional Phrase, Unnecessary Adjective, Unnecessary Adverb, Unnecessary Auxiliary Verb, Unnecessary Conjunction, Unnecessary Particle, Unnecessary Noun, Unnecessary Pronoun, Spelling Error & 5 \\
    \bottomrule
    \end{tabularx}
    \caption{The grammar error hierarchy that we employ in our system.}
    \label{tab:grammar_hierarchy}
\end{table}

\section{Conversation Data Synthesis}

\subsection{Topics}
\label{app:convo_topics}

Table~\ref{tab:num_topics} details the number of topics per broad topic area. For a complete list of topics, please see Appendix~\ref{app:complete_topics}.

\begin{table}[!h]
    \centering
    \begin{tabular}{cc}
    \toprule
    \textbf{Topic Area} & \textbf{Topic Counts}\\
    \midrule
    Food & 36 \\
    Books & 43 \\
    Movies & 44 \\
    TV shows & 31 \\
    Music & 45 \\
    Hobbies & 34 \\
    English learning & 10\\ 
    \textbf{Total} & 243\\
    \bottomrule
    \end{tabular}
    \caption{Number of topics per topic area.}
    \label{tab:num_topics}
\end{table}

% \includepdf[pages=-]{latex/Topics.pdf}

\subsection{Generation Diversity Issues and Corresponding Prompt Adjustments}
\label{app:gen_div_issues}
In \citet{li2023curriculum}, ChatGPT is first prompted to generate two distinct personas, one generic persona (Person 1, often assumed to be American by ChatGPT), and one Chinese college student persona (Person 2). The LLM is then asked to generate a conversation using these personas, where Person 1 should lead the conversation by asking questions and sharing engaging anecdotes when appropriate. Multiple conversations with various persona choices are generated for each textbook topic and corresponding vocabulary set.

We first adjust the requirements of the conversation generation step to make Person 1 more empathetic and attentive to Person 2. We also remove the vocabulary constraint and request that the generated conversations be spoken. However, we notice an alarming homogeneity in the generated conversations in terms of Person 2's preferences. For example, out of seven conversations generated about favorite foods, three feature hot pot, three feature dumplings, and one features Peking Duck. For favorite songs, "The Moon Represents My Heart" is Person 2's favorite in six out of ten conversations, and generic old Chinese songs are favorites in the other conversations. These overly repetitive examples do not represent the general population of Chinese college students.
As a result, to enhance the diversity of our training data, we define Person 2 as someone whose first language is not English.

\subsection{Data Synthesis Prompts}
\label{app:synth_prompts}

The following is the prompt used to generate different personas:
\begin{lstlisting}
personas_prompt = ("Please provide me with one individual Person 1 with different backgrounds, "
               "including information about their demographic, socio-economic status, culture, MBTI personality type, and personal experiences, "
               "no need to show names. "
               "Then provide me with one individual Person 2 who is a college student but with different information; Person 2's native language is not English.")
\end{lstlisting}

After generating the personas, given a specific topic, we use the following prompt to generate 10 conversations about this topic within the same ChatGPT prompting session:

\begin{lstlisting}
convo_prompt = (
    'Generate a single spoken conversation between these two people as Person 1 and Person 2 about the topic "{topic}".\n'
    "Please take into account their distinct personalities and their backgrounds. Begin the conversation with Person 1.\n"
    "Person 1 should guide the conversation by asking more questions; Person 1 should also be attentive to Person 2's interests and ask Person 2 to say more.\n"
    "Person 1 should be able to make specific recommendations to Person 2 if requested. Person 2 should feel free to ask for recommendations from Person 1 if appropriate.\n"
    "Begin the conversation with Person 1. Person 1 does not know any information about Person 2 unless Person 2 brings it up. Person 1 should not recommend restaurants, stores, or recipes. Keep utterances colloquial. Person 1 should discuss the recommendation directly in conversation, rather than saying they will send the recommendations later. The conversation should last at least 10 turns.")
\end{lstlisting}

\subsection{Data Filtering Mechanism}
\label{app:data_filter}

We apply the same data formatting filtering as \citet{li2023curriculum} (e.g. making sure the conversation starts with Person 1, etc). We additionally include filtering mechanisms for our specific issues in dialogue generation. Specifically:

\begin{enumerate}
    \item Person 1 would make assumptions about Person 2 without Person 2 mentioning it (e.g. asking Person 2 whether they miss Brazil even though Person 2 has not mentioned that they are Brazilian). This is likely because ChatGPT assumes Person 1 knows Person 2's persona. 
    \item Person 1 would offer to send their recommendations via private message. While this is likely in everyday conversation, since our chatbot does not have a mechanism for private messages, this is considered a failure mode.
\end{enumerate}

We now present our filtering mechanisms. Given a dialogue history string, we provide ChatGPT with the following prompt:
\begin{lstlisting}
"Does Person 1 in the following conversation make assumptions about Person 2 without the user bringing it up first? Answer with yes or no.\n\n" + dialogue_string
\end{lstlisting}

If the ChatGPT response starts with "yes", then we filter out this dialogue.

If not, we pass the conversation through one more layer of filtering:
\begin{lstlisting}
"Does Person 1 in the following conversation make specific recommendations when requested? If Person 2 does not request specific recommendations, answer \"Yes\". Answer with yes or no.\n\n" + dialogue_string
\end{lstlisting}

If the ChatGPT response starts with "no", then this conversation is pruned.

\subsection{Topic Distribution over Conversations}
\label{app:topic_dist}

% {'food': 124, 'hobbies': 195, 'movies': 209, 'music': 233, 'books': 243, 'tv shows': 167, 'english learning': 56}
\begin{table}[!h]
    \centering
    \begin{tabular}{cc}
    \toprule
    \textbf{Topic} & \textbf{Conversation Counts}\\
    \midrule
    Food & 124 \\
    Books & 243 \\
    Movies & 209 \\
    TV shows & 167 \\
    Music & 233 \\
    Hobbies & 195\\
    English learning & 56 \\ 
    \textbf{Total} & 1227 \\
    \bottomrule
    \end{tabular}
    \caption{Number of conversations per topic after pruning}
    \label{tab:my_label}
\end{table}

\section{Personalization Details}
\label{app:personalization}

\subsection{Short and Succinct Feedback}
When the participant selects that they prefer succinct feedback and their experimental condition allows for feedback personalization, we use the following prompt along with the past three student utterances (\texttt{convo}) and the original generated feedback (\texttt{output}):
\begin{lstlisting}
    Given the following utterances by a student learning English as the context:\n\n{convo}\n\nAnd a piece of feedback:\n\n{output}\n\nMake it more succinct and concise while retaining the original examples with their full context. Make the feedback colloquial and succinct. Don't use the word \"basic\". Try to shorten to at most 3 sentences.
\end{lstlisting}

\subsection{Detailed Feedback with Examples}
When the participant prefers their feedback to have more detail, we use the following sequence of prompts with the past three student utterances (\texttt{convo}) and the original generated feedback (\texttt{output}). We use a sequence of prompts instead of a single prompt because we notice that ChatGPT often overgenerates on the detail and makes the feedback too long. 

\begin{lstlisting}
    Given the following utterances by a student learning English as the context:\n\n{convo}\n\nAnd a piece of feedback:\n\n{output}\n\nCreate a new piece of feedback with more context-specific examples supporting the feedback. Make the feedback colloquial, as if spoken in conversation. Don't use the word \"basic\".
\end{lstlisting}

Upon obtaining the ChatGPT generation, we use the following prompt to shorten the feedback:
\begin{lstlisting}
    Shorten your response to 3 - 4 sentences while retaining necessary information and detail.
\end{lstlisting}

\section{Transition Improvements}
\label{app:transitions}

\subsection{Overview}
After receiving a piece of feedback, the user may have some questions about the feedback that are directly related to English learning. If that is the case, our conversation model may not be able to handle them well since they are better trained on open-domain chit-chat. Therefore, we should ensure to use ChatGPT to handle these queries.

We classify whether a user utterance is a relevant query using ChatGPT. If the query is directly related to English learning, ChatGPT is instructed to produce a response. This response is presented to the user directly without any modification. If the user utterance is not a query or the query is not directly related to English learning, the ChatGPT response is processed such that all questions are removed from the response. This altered response is then concatenated with a connector phrase with a recap of the conversation before the feedback, as well as the response from the conversation model prior to activating the feedback module.

\subsection{Query Classification}

Given a conversation history of the last three turns, we classify the query using the following ChatGPT prompt:
\begin{lstlisting}
    Given the following user-chatbot exchange:\n\n{convo_history}\n\nIs the latest user utterance asking for clarifications or English learning advice? Answer with yes or no.
\end{lstlisting}

\subsection{ChatGPT Response Processing}

We generate the response to the user utterance using this ChatGPT prompt:

\begin{lstlisting}
    Respond to the last user utterance as the Assistant based on the conversation context. Be colloquial and helpful. You only know English and Mandarin.
\end{lstlisting}

If the latest user utterance is not a query or a relevant query, we process the ChatGPT response to remove any questions. Specifically, we first tokenize the response into individual sentences and concatenate sentences that do not end with question marks together. We then use ChatGPT to very briefly summarize the pre-feedback conversation using this prompt:

\begin{lstlisting}
    Given the following conversation history:\n\n{convo}\n\nDescribe the current general topic with ONE SHORT PHRASE.
\end{lstlisting}

We then create a connector sentence employing a pre-defined set of connector phrases. \texttt{curr\_topic} here refers to the conversation summary phrase.

\begin{lstlisting}
    f"Alright, let's continue our conversation about {curr_topic}.", f"Let's get back to our chat on {curr_topic}!",
    f"Okay let's go back to our conversation about {curr_topic}.", f"Now back to our conversation with respect to {curr_topic}.",
    f"Lets' go back to our chat. We just talked about {curr_topic}.", f"Let's keep chatting about {curr_topic}.",
    "Okay, let's keep chatting.", "Let's go back to our conversation!", "Let's continue our chat!"
\end{lstlisting}

\section{Fixed Empathetic Responses}
\label{app:fixed_emp}

The full list of fixed empathetic response utterances is as follows:
\begin{enumerate}
    \item I understand that learning English can be a difficult process, but you are doing great! Keep it up!
    \item I have been seeing steady progress from you. English learning can be challenging, but you are doing so well!
    \item I'm really proud of the progress you're making. Keep powering through it and you will see even more significant improvement in your spoken English!
    \item Your hard work on learning English is really paying off. You have been nailing this conversation thus far!
    \item I can see how much you've improved in your spoken English. Keep it up and you will see even more improvement in no time!
    \item Keep up the great work in practicing your spoken English; every effort brings you a step closer to fluency.
\end{enumerate}

\section{Further Experiment Details}
\subsection{Informed Consent}
We present our informed consent form through this link: \url{https://github.com/siyan-sylvia-li/EDEN/blob/main/experimental_data/Informed%20Consent%20-%20Public%20-%20Google%20Forms.pdf}. 

\subsection{Pre- and Post-surveys}
\label{app:pre_post_survey}

In the pre-survey, in addition to the L2 grit survey, the participants are asked to complete the following questions about their English proficiency:

\begin{enumerate}
    \item How many years have you been learning English?
    \item How is your English proficiency? (1 is beginner and 5 is proficient)
    \item What was your IELTS score?
    \item What was your TOEFL score?
    \item How much English do you speak every day? (1 is Mandarin only, 5 is English only)
\end{enumerate}

In the post-survey, the participants are first asked to complete the questions about general conversation quality, then the questions about chatbot PAS, and then, finally, the L2 grit survey.

% \includepdf[pages=-]{latex/informed_consent_blackout.pdf}

\section{Example Conversations}
\label{app:example_convo}

In the following examples, we use {\color{red}Red} texts to indicate adaptive empathetic feedback,  {\color{blue}Blue} texts to indicate fixed empathetic feedback, and {\color{orange}Orange} texts to indicate grammatical feedback.

\subsection{Adaptive Empathetic Feedback Dialogue Example}

\noindent \textbf{USER:} No, I Sorry, I normally just just How should I describe Normally when I cooking I just follow in my mind and you know I'm from China. I know the recipe taste like and I know what kind of the sauce I should put in when I cooking

\noindent \textbf{BOT:} {\color{red}You're doing well with the topic, but work on your grammar to sound smoother. For instance, say "I know what the recipe should taste like" instead of "I know the recipe taste like." Keep practicing to get even better!  Does that sound alright to you?}

\noindent \textbf{USER:} Yes, thank you. Thank you for the instruction. And we move on to the next topic.

\noindent \textbf{BOT:} Great! If you're ready, let's jump into the next topic. Just let me know if you need any help or have any questions along the way. Okay let's go back to our conversation about recipe sources. Ah, that makes a lot of sense. It's wonderful that you have such a strong connection to your culture and food. It's like a bridge to your roots, right? I can imagine that cooking must bring back a lot of memories and emotions. Have you ever tried to share some of your Chinese dishes with friends or family here?

\noindent \textbf{USER:} Yeah, definitely. When we have a party with my friends, Chinese friends, yeah. Like I said before, we're cooking together and enjoy the meal together. We even can discuss which one is authentic, which one... What's that preference?

\noindent \textbf{BOT:} It's great that you're so thoughtful about the authenticity. It's important to have those discussions, especially when sharing something so meaningful. I think it's a beautiful way to connect with others. Speaking of which, I recently had a similar experience with a Peruvian dish. It was interesting to see how others perceived it and how it was passed down.

\noindent \textbf{USER:} Yeah, that can make a topic when we hang out together. So people can share in the life experience and we can have more topics, more subjects to talk.

\noindent \textbf{BOT:} {\color{orange}Maybe you meant "So people can share in their life experiences and we can have more topics and subjects to talk about.".   Does that sound alright to you?}

\begin{figure*}[!ht]
    \centering
    \includegraphics{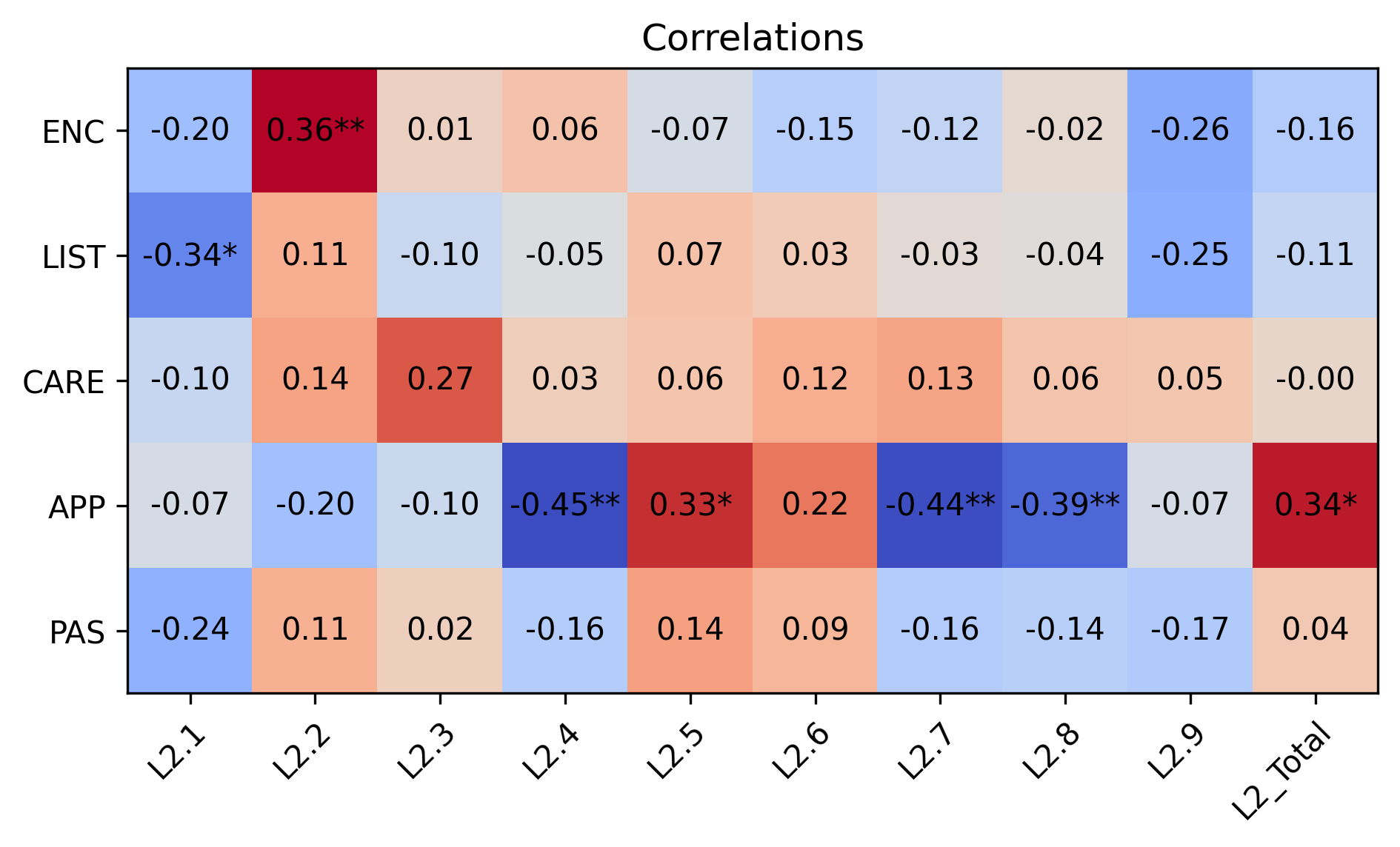}
    \caption{Correlations between different measures for chatbot PAS and L2 grit. **$ = p < 0.05$, *$ = p < 0.1$}
    \label{fig:pas_grit_plot}
\end{figure*}

\subsection{Fixed Empathetic Feedback Dialogue Example}
\noindent \textbf{USER:} Recently I'm watching Friends, that's a really classic sitcom but I'm really enjoying that. Now I'm coming to the sixth episode. So do you like this sitcoms and how do you find the characters?

\noindent \textbf{BOT:} Friends is a great choice! I've watched it a few times, and I can see why it's so popular. The characters are definitely unique and relatable in their own ways. How do you find the comedy in the show, and do you have a favorite character?

\noindent \textbf{USER:} That's an interesting question. Before I liked Ross because I think he's really reliable and I like smart people but now I gradually like Chandler more because I found he's really you know the type of dream partner I would say. He treats Monica really well and he's funny by the way yeah I like humorous people

\noindent \textbf{BOT:} {\color{blue}I have been seeing steady progress from you. English learning can be challenging, but you are doing so well!}

\noindent \textbf{USER:} Thank you. Actually, I still think I lack competence and also the fluency in my speaking English. Oftentimes, I don't know how to express my feeling quite well and also it's difficult to understand the idioms, some common expressions. So do you have any suggestions according to that, like how to improve my daily English speaking?

\noindent \textbf{BOT:} No worries at all! Improving your English speaking skills is totally doable. One suggestion is to practice speaking with someone regularly, maybe a classmate or a language exchange partner. You could also try watching English movies with subtitles to pick up on common expressions and idioms. Remember, practice makes perfect, so keep at it! Good luck!

\noindent \textbf{USER:} Thank you for saying that. Maybe I want to improve my pronunciation because I often feel my speaking is not authentic as a native speaker and you can easily tell that I'm a foreigner. I really want to improve that part. Do you have any suggestions?

\noindent \textbf{BOT:} No worries at all! Improving your pronunciation is a great goal. One tip is to practice speaking slowly and focus on the sounds of each word. You could also try mimicking native speakers or even recording yourself to hear how you sound. Keep at it, and don't be too hard on yourself - progress takes time! Good luck! Let's go back to our conversation! I can see why Chandler's character appeals to you. His wit and humor really make the show more enjoyable. It's great to hear that you're enjoying the show so far. Do you think the cultural differences between the characters add to the humor or do you find that they sometimes create challenges in understanding their dynamics?

\subsection{Grammatical Feedback Dialogue Example}
\noindent \textbf{USER:} For that part, I do agree it has brought positive impact onto the whole entertainment business with more representations of Asian experiences.

\noindent \textbf{BOT:} {\color{orange}I believe you wanted to say "For that part, I do agree it has brought a positive impact onto the whole entertainment business by representing more Asian experiences." Does that sound good?}

\noindent \textbf{USER:} Oh yeah, that sounds good.

\section{Additional Details for Results}
\subsection{PAS and Conversation Quality}
\label{app:pas_conv}

We analyze the relationship between overall PAS and our different measures for conversation quality, as well as the relationship between overall L2 grit changes and our measures for conversation quality. We present the result of our correlational analysis in Table \ref{tab:pas_l2_conv}.

\begin{table}[!h]
    \centering
    \begin{tabular}{llccc}
    \toprule
        \textbf{Measure} & & \textbf{QUAL} & \textbf{CONF} & \textbf{USE} \\
    \midrule
    PAS & Coef. & 0.28& 0.54 & 0.62 \\
    & p-value & 0.13 & 0.0018 & 0.0002\\
    \midrule
    $\Delta$L2\_Total & Coef. & -0.17& -0.19 & 0.05 \\
    & p-value & 0.35& 0.31& 0.78\\
    \bottomrule
    \end{tabular}
    \caption{Correlations between PAS and different measures for conversational quality, as well as correlations between L2 grit and measures for conversational quality.}
    \label{tab:pas_l2_conv}
\end{table}

\subsection{PAS and L2 Grit}
\label{app:pas_grit}

Here, we analyze the relationship between PAS measures and L2 grit, rather than changes in L2 grit. We present the resulting correlation matrix in Figure \ref{fig:pas_grit_plot}. We see that APP has negative correlations with the reverse-coded items in the L2 grit scale, and is positively correlated with total L2 grit in the post-survey. We see that LIST also has a weak negative correlation with L2.1, similar to how CARE correlates negatively to $\Delta$L2.1.

\section{Complete Topics List}
\label{app:complete_topics}
\begin{lstlisting}
FOOD
['Cooking traditions in family gatherings', 'Exploring cultural significance through food memories', 'Nostalgic meals from childhood', 'Evolution of taste preferences over time', 'Food-related rituals and celebrations', 'Culinary adventures while traveling', 'Impact of favorite food-related memories on overall well-being', 'Favorite food', 'Cultural significance of favorite foods', 'Psychological aspects of comfort foods', 'Historical origins of popular dishes', 'Regional variations in favorite foods', 'Impact of advertising on food choices', 'Fusion cuisine and blending of flavors', 'Favorite cuisine', 'Fusion cuisines incorporating favorite elements', 'Health benefits of favorite cuisines', 'Popular street foods within favorite cuisines', 'Vegan/vegetarian adaptations of favorite cuisines', 'Cultural significance of ingredients in favorite cuisines', 'Cooking techniques specific to favorite cuisines', 'Famous chefs and restaurants specializing in favorite cuisines', 'Street food preferences', 'Global street food culture', 'Health considerations in street food', 'Popular street food vendors around the world', 'Street food festivals and events', 'DIY street food recipes', 'Historical evolution of street food', 'Street food and cultural identity', 'Sustainable practices in street food markets', 'Street food safety regulations', 'Street food fusion trends', 'The Role of Food in Celebrations and Festivals', 'Cultural significance of traditional dishes in festivals', 'Evolution of festival foods over time']

HOBBIES
['Finding time for hobbies', 'Time management techniques', 'Exploring leisure activities', 'Prioritizing personal interests', 'Balancing work and leisure', 'Creating a hobby schedule', 'Discovering passion projects', 'Incorporating relaxation into daily routine', 'Maximizing free time', 'Setting goals for hobbies', 'Joining hobby groups or clubs', 'What to do during free time', 'Hobbies to Pursue', 'Outdoor Activities to Try', 'Creative Projects to Start', 'Indoor Activities for Relaxation', 'DIY Projects to Explore', 'Social Activities to Engage In', 'Learning New Skills', 'Volunteering Opportunities', 'Cultural Events to Attend', 'Wellness Practices for Self-care', 'New hobbies', 'Picking up new hobbies', 'Outdoor activities', 'Crafting and DIY projects', 'Gardening and urban farming', 'Cooking and baking', 'Fitness and exercise routines', 'Music production and learning instruments', 'Painting and drawing', 'Photography and videography', 'Creative writing and journaling', 'Board games and tabletop gaming']

MOVIES
['Favorite movie', 'Movie genres and their characteristics', 'Impact of favorite movies on personal taste', 'Analysis of favorite movie soundtracks', 'Cultural significance of favorite movies', 'Evolution of movie preferences over time', 'Favorite movie directors and their filmography', 'Psychology behind attachment to favorite movies', 'Societal influence on favorite movie choices', 'Comparing favorite movies with critical acclaim', 'The role of nostalgia in favorite movie selection', 'Favorite movie director', 'Filmography analysis of favorite movie director', 'Influence of favorite movie director on modern cinema', 'Cinematic style of favorite movie director', 'Collaborations with actors/actresses by favorite movie director', "Favorite movie director's impact on the industry", "Favorite movie director's signature themes and motifs", "Evolution of favorite movie director's directing techniques", "Comparison of favorite movie director's works with contemporaries", "Behind-the-scenes insights into favorite movie director's creative process", 'Legacy of favorite movie director in film history', 'Favorite movie genre', 'Action-packed films', 'Romantic comedies', 'Sci-fi and fantasy flicks', 'Horror movies', 'Historical dramas', 'Animated features', 'Mystery and thriller genres', 'Documentaries', 'Musical films', 'Adventure movies', 'What makes a good movie', 'Character development in films', 'Plot structure and storytelling techniques', 'Visual aesthetics and cinematography', 'Soundtrack and musical score impact', 'Effective use of symbolism and motifs', 'Genre conventions and audience expectations', 'Impact of pacing and editing on viewer engagement', 'Dialogue and scriptwriting excellence', 'Cultural and societal influences on film reception', 'Directorial style and vision manifestation']

MUSIC
['Favorite song', 'Music genres', 'Lyric analysis', 'Musical composition techniques', 'Influence of culture on music preferences', 'Evolution of music over decades', 'Impact of technology on music production', 'Music therapy benefits', 'Famous songwriters and their work', 'Music and emotions', 'Role of music in society', 'Favorite musical artist', 'Favorite band', 'Favorite musical genre', 'History of jazz music', 'Evolution of rock and roll', 'Impact of hip hop culture', 'Classical music composers', 'Folk music traditions around the world', 'Influence of electronic music on modern culture', 'Pop music trends and analysis', 'Traditional music instruments of various cultures', 'Fusion genres in contemporary music', 'Music therapy and its benefits', 'How music makes you feel', 'Psychological effects of music', 'Emotional impact of music', 'Music therapy benefits', 'Neuroscience of music and emotions', 'Music and mood regulation', 'Cultural influences on music perception', 'Music and memory recall', 'Physiological responses to music', 'Music and stress reduction', 'Social bonding through music', 'Playing musical instruments', 'Music theory', 'Learning techniques', 'Instrument maintenance', 'Historical development of instruments', 'Musical genres', 'Famous musicians', 'Music composition', 'Instrument accessories', 'Performance techniques']

TV SHOWS
['Favorite TV Show', 'Character Development in TV Shows', 'Impact of TV Shows on Culture', 'Evolution of TV Show Genres', 'Representation in Television', 'Exploring TV Show Soundtracks', 'The Role of Television in Storytelling', 'Favorite TV character', 'Character development in TV shows', 'Impact of TV characters on audience', 'Evolution of TV show protagonists', 'Analysis of popular TV show archetypes', 'Gender representation in TV show characters', 'Cultural significance of iconic TV characters', 'Character arcs in long-running TV series', 'Favorite TV genre', 'Favorite Comedy Series', 'Favorite Drama Series', 'Favorite Crime Shows', 'Favorite Science Fiction Series', 'Favorite Fantasy Series', 'Favorite Documentary Series', 'Favorite Reality TV', 'Favorite Animated Series', 'Favorite Historical Drama', 'Favorite Thriller Series', 'TV show binge-watching habits', 'Streaming platforms usage', 'Effects of binge-watching on sleep', 'Psychological impact of binge-watching', 'TV show reboots and revivals']

BOOKS
['Favorite book', 'Favorite novel', 'Favorite non-fiction', 'Favorite fiction', 'Favorite author', 'Favorite authors', 'Favorite book genres', 'Literary influences', 'Writing styles', 'Character development techniques', 'Plot structures', 'Narrative perspectives', 'Symbolism in literature', 'Authorial voice', 'Classic literature', 'Literary analysis techniques', 'Historical context in literature', 'Themes in classic literature', 'Famous authors of classic literature', 'Impact of classic literature on society', 'Gender roles in classic literature', 'Adaptations of classic literature in film and theater', 'Book clubs', 'Reading habits', 'Community engagement through books', 'Social impact of book clubs', 'Diversity in reading selections', 'Virtual book club trends', 'Must-read books', 'Classic literature books', 'Modern fiction books', 'Non-fiction bestsellers', 'Biographies and memoirs', 'Science fiction and fantasy books', 'Self-help and personal development books', 'History and politics books', 'Philosophy and spirituality books', 'Crime and mystery books', 'Young adult literature books', 'Bookstores and libraries', 'Audiobooks vs. physical books', 'Book adaptations (movies, TV shows, etc.)', 'Fiction vs. non-fiction']

ENGLISH LEARNING
['Vocabulary acquisition and expansion', 'Grammar rules and structures', 'Pronunciation practice', 'Reading comprehension strategies', 'Writing skills development', 'Listening comprehension exercises', 'Speaking fluency and conversation practice', 'Idioms and expressions', 'Cultural aspects and context in English language learning', 'Test preparation (e.g., TOEFL, IELTS, Cambridge exams)']



\end{lstlisting}

\end{document}